\documentclass[sigconf]{acmart}

\AtBeginDocument{%
  }

\setcopyright{acmlicensed}
\copyrightyear{2018}
\acmYear{2018}
\acmDOI{XXXXXXX.XXXXXXX}

\acmConference[Conference acronym 'XX]{Make sure to enter the correct
  conference title from your rights confirmation emai}{June 03--05,
  2018}{Woodstock, NY}
\acmISBN{978-1-4503-XXXX-X/18/06}

\usepackage{amsmath,amssymb,amsfonts,bm}
\usepackage{url}
\usepackage{multirow}
\usepackage{subcaption}
\usepackage[linesnumbered,vlined,ruled,noend]{algorithm2e}
\usepackage{graphicx}  

\begin{document}

\title{Physics-guided Active Sample Reweighting for Urban Flow Prediction}
\author{Wei Jiang}
\affiliation{%
  \institution{The University of Queensland}
  \city{Brisbane}
  \country{Australia}
}
\email{wei.jiang@uq.edu.au}

\author{Tong Chen}
\affiliation{%
  \institution{The University of Queensland}
  \city{Brisbane}
  \country{Australia}}
\email{tong.chen@uq.edu.au}

\author{Guanhua Ye}
\affiliation{%
  \institution{Beijing University of Posts and Telecommunications}
  \city{Beijing}
  \country{China}}
\email{g.ye@bupt.edu.cn}

\author{Wentao Zhang}
\affiliation{%
  \institution{Peking University}
  \city{Beijing}
  \country{China}}
\email{wentao.zhang@pku.edu.cn}

\author{Lizhen Cui}
\affiliation{%
  \institution{Shandong University}
  \city{Jinan}
  \country{China}}
\email{wentao.zhang@pku.edu.cn}

\author{Zi Huang}
\affiliation{%
  \institution{The University of Queensland}
  \city{Brisbane}
  \country{Australia}}
\email{huang@itee.uq.edu.au}

\author{Hongzhi Yin}
\authornote{Corresponding author.}
\affiliation{%
  \institution{The University of Queensland}
  \city{Brisbane}
  \country{Australia}}
\email{h.yin1@uq.edu.au}

\renewcommand{\shortauthors}{Wei Jiang et al.}

\begin{abstract}
Urban flow prediction is a spatio-temporal modelling task that estimates the throughput of transportation services like buses, taxis, and ride-sharing, where data-driven models have become the most popular solution in the past decade. Meanwhile, the implicitly learned mapping between historical observations to the prediction targets tend to over-simplify the dynamics of real-world urban flows, leading to suboptimal predictions. Some recent spatio-temporal prediction solutions bring remedies with the notion of physics-guided machine learning (PGML), which describes spatio-temporal data with nuanced and principled physics laws, thus enhancing both the prediction accuracy and interpretability. However, these spatio-temporal PGML methods are built upon a strong assumption that the observed data fully conforms to the differential equations that define the physical system, which can quickly become ill-posed in urban flow prediction tasks. The observed urban flow data, especially when sliced into time-dependent snapshots to facilitate predictions, is typically incomplete and sparse, and prone to inherent noise incurred in the collection process (e.g., uncalibrated traffic sensors). As a result, such physical inconsistency between the data and PGML model significantly limits the predictive power and robustness of the solution. Moreover, due to the interval-based predictions and intermittent nature of data filing (e.g., one record per 30 minutes) in many transportation services, the instantaneous dynamics of urban flows can hardly be captured, rendering differential equation-based continuous modelling a loose fit for this setting. To overcome the challenges, we develop a discretized physics-guided network (PN), and propose a data-aware framework \underline{P}hysics-\underline{g}uided \underline{A}ctive \underline{S}ample \underline{R}eweighting (P-GASR) to enhance PN. Technically, P-GASR incorporates an active sample reweighting pipeline, which not only minimizes the model uncertainty of PN to enhance robustness, but also prioritizes data samples that exhibit higher physical compliance to reinforce their contribution to PN training. Experimental results in four real-world datasets demonstrate that our method achieves state-of-the-art performance with a demonstrable improvement in robustness. The code is
released at \url{https://github.com/WeiJiang01/P-GASR}.
\end{abstract}

\ccsdesc[500]{Information systems~Spatial-temporal systems}

\keywords{Physics-guided AI; Urban Flow Prediction; Spatio-Temporal Data Mining; Sample Reweighting}
\copyrightyear{2024}
\acmYear{2024}
\setcopyright{acmlicensed}\acmConference[CIKM '24] {Proceedings of the 33rd ACM International Conference on Information and Knowledge Management}{October 21--25, 2024}{Boise, ID, USA.}
\acmBooktitle{Proceedings of the 33rd ACM International Conference on Information and Knowledge Management (CIKM '24), October 21--25, 2024, Boise, ID, USA}
\acmDOI{10.1145/3627673.3679738}
\acmISBN{979-8-4007-0436-9/24/10}

\maketitle
\section{Introduction}

With the ongoing development of modern cities, the demand for urban spatio-temporal modelling applications is rising. Accurate urban spatio-temporal prediction and modelling (e.g., traffic prediction and passenger demand prediction) are crucial for assessing and mitigating various trends and events, as well as for effective urban planning. In recent years, the primary solutions for urban spatio-temporal prediction have leveraged data-driven models, which have demonstrated remarkable achievements \cite{tedjopurnomo2020survey, yuan2021survey, yin2016spatio, wang2017location}. A general paradigm for modelling spatio-temporal urban flow consists of constructing graphs from geographical grids or sensor networks and utilizing spatio-temporal graph neural networks to capture location- and time-dependent information simultaneously \cite{yu2017spatio, li2021spatial, mohamed2020social, zhang2020spatio, song2020spatial, yin2015joint}. Moreover, to further enhance the capability of these models, some of the studies integrate self-supervised learning as an auxiliary task to improve the model's robustness through data augmentation \cite{ji2023spatio, qu2022forecasting, zhou2021self, yin2016spatio}.

Despite data-driven models showing promising applicability to urban flow prediction, long-term accurate prediction requires extensive and high-quality data to facilitate training, which is not always available. On the other hand, using traditional physics-based models directly for prediction bypasses the need for a huge data pool but unfortunately faces several limitations. For instance, formulating a complex physical system for spatial-temporal dynamic modelling inevitably suffers from low computational efficiency and is time-consuming \cite{seyyedi2023machine, willard2020integrating}. Additionally, physical systems' performance is highly dependent on the calibration of their parameters (i.e., coefficients in a physics formula), where inaccurate parameterizations can lead to model biases and imprecise predictions \cite{sulis2010comparison,wang2021physics}. As a remedy, many studies instead introduce physics-guided machine learning (PGML) approaches \cite{rodriguez2023einns,he2023physics,ji2022stden,hettige2024airphynet,chen2023physics,chen2022physics,jia2021physics}. These methods integrate physics knowledge as constraints for machine learning models, allowing them to capture spatio-temporal dynamics while balancing data efficiency and model generalizability. Within these PGML approaches, some methods augment data by generating simulated data from physics-based models \cite{chen2023physics,chen2022physics,jia2021physics}, and some other methods improve model performance by either incorporating differential equations into the graph convolutional structure \cite{ji2022stden,hettige2024airphynet}, or integrating physical constraints into the loss function to correct the direction of optimization \cite{rodriguez2023einns,he2023physics}.

\begin{figure}[t]
    \centering
    \begin{minipage}[t]{0.85\linewidth}
    \centering
    \includegraphics[width=\linewidth]{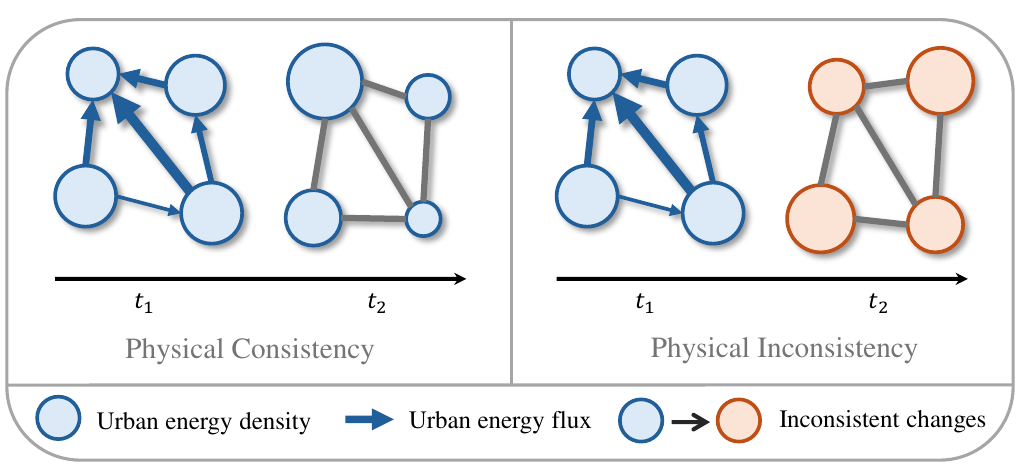}
    \subcaption{A toy example illustrates that urban energy density changes over time in accordance with conservation law when subject to physical consistency, whereas physical inconsistency results in a violation of the conservation law. The arrows represent the direction of urban energy flux flow, the thickness of the edges represents the magnitude of the flux, and the size of the nodes reflects the urban energy density.}
    \label{fig:problem_illustration}
    \end{minipage}
    \begin{minipage}[t]{0.85\linewidth}
    \vspace{0.3cm}
    \centering
    \includegraphics[width=\linewidth]
    {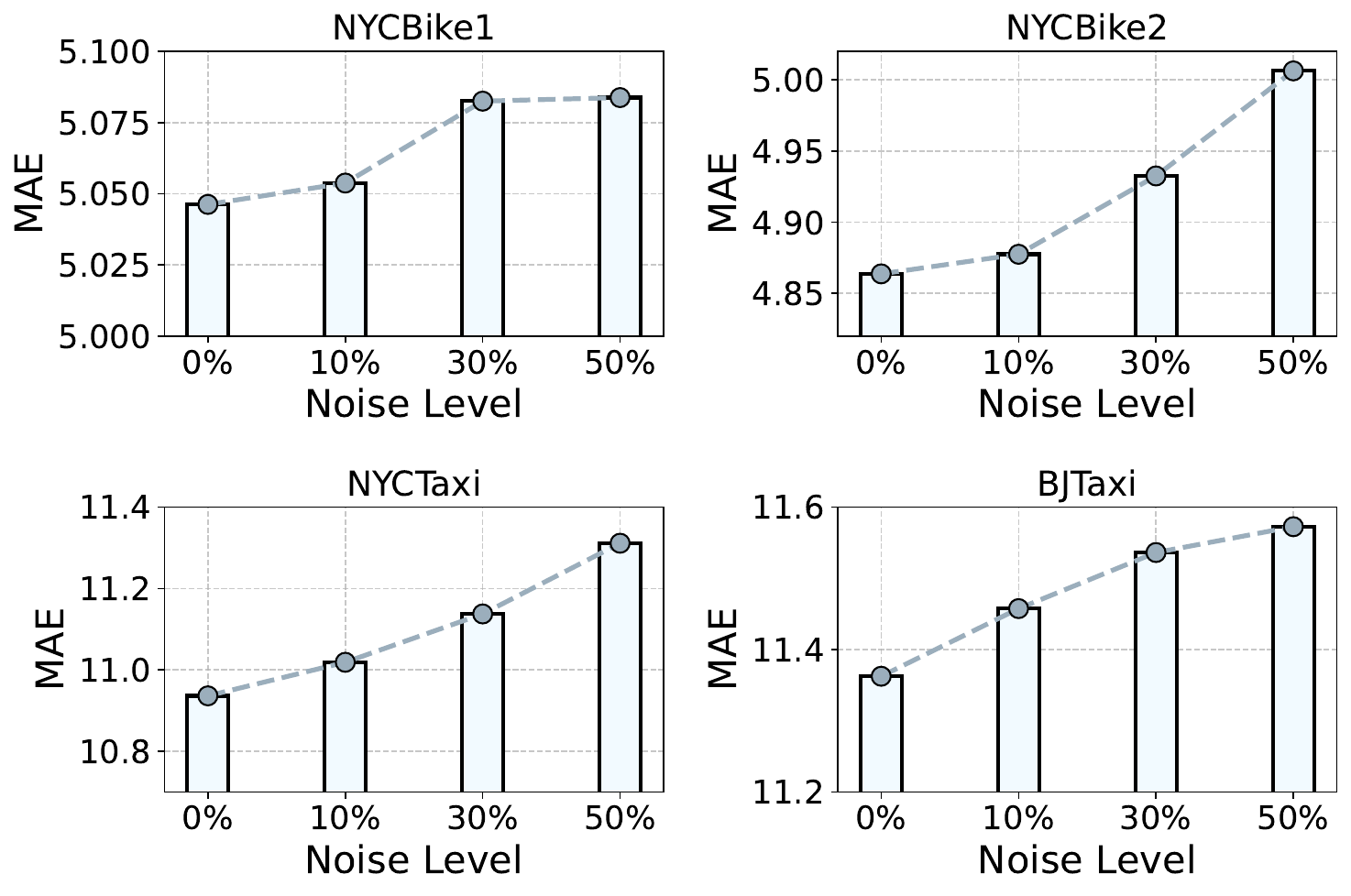}
    \subcaption{Influence of different noise levels on physics-guided network on four passenger demand prediction datasets (see experimental settings in Sec. \ref{sec:experiment_set} and \ref{sec:noise_influence}).}
    \label{fig:problem_illustration_noise}
    \end{minipage}
    \caption{Physical inconsistency in urban flow network based on an example and data observation. }
    \label{fig:illustration}
    \vspace{-3mm}

    \end{figure}

 In short, when addressing data scarcity that challenges data-driven spatio-temporal models, PGML is an effective solution with higher efficiency and flexibility compared with traditional physics-based methods. A fundamental assumption underpinning all PGML methods is that the observed data and corresponding physical knowledge are consistent. However, this is not necessarily the case in urban flow prediction tasks, where the real-world observational data may not conform to established physical laws. This issue can be attributed to limited data availability, sensor errors, or discrepancies between the physical systems and real-world scenarios. Incorporating physical knowledge into models when there is a mismatch between data and physical principles can potentially mislead the model, creating bottlenecks for achieving optimal performance. For instance, Fig. \ref{fig:problem_illustration} shows an example of physical inconsistency, which indicates that the input data do not conform to physical laws (see definitions in Sec. \ref{sec:definiton}). By intuition, PGML is an effective improvement over traditional data-driven approaches, but this is limited to situations where data quality is assured. As a proof-of-concept, we conducted a preliminary experiment using our proposed physics-guided model (see details in Sec. \ref{sec:pe}). By injecting Gaussian noise into a certain proportion of data points, we simulate different levels of physical inconsistency in the training data. As can be seen from Fig. \ref{fig:problem_illustration_noise}, as the noise level increases, the performance of PN progressively worsens. Even 10-30\% of noisy samples can lead to significant performance degradation in PGML methods, demonstrating the detrimental effects of inconsistencies between data and physical knowledge. This highlights the importance of ensuring consistency between the data used and the physical principle applied in PGML methods to avoid undermining the effectiveness of the prediction. Given that it is impractical to guarantee that the urban flow data is noise-free, it becomes desirable to instead make a PGML model more robust to such noises.


In light of the inherent noises within the data, we aim to maximize the potential of a PGML model for urban flow prediction by uplifting its awareness of data quality. However, straightforwardly using data selection methods like core sets is a less feasible solution in the context of PGML, as the utility of each data sample is additionally correlated to the physics component and is subject to different stages during training. Thus, we adopt a novel, adaptive reweighting approach that jointly considers two subproblems: (i) identifying data points that are beneficial for the overall risk minimization - which aligns with the default objective of existing sample reweighting mechanisms \cite{ren2018learning}; and (ii) filtering out data points that do not comply with the embedded physics in the PGML model to ensure the urban dynamics are correctly captured. To this end, we propose a sample reweighting framework named Physics-guided Active Sample Reweighting (P-GASR), which can lower the impact from physically consistent and inconsistent samples to the PGML model to be learned. Technically, an active reweighting policy is introduced to determine a sample's weight by simultaneously measuring its contributions to model generalization via model uncertainty quantification, and to learning meaningful physical information with the discrete conservation law in urban flow network \cite{bressan2015conservation}. Guided by the sample weights, the PGML model can selectively favour samples that exert high model uncertainty and exhibit high physical consistency during the training process. 

Moreover, it is also worth noting that, the reweighting pipeline in P-GASR is designed in an active learning fashion. Unlike other sample reweighting methods that take an extra meta-learning step to learn each sample's weight \cite{ren2018learning}, our active reweighting policy computes sample weights directly in the forward pass. This eliminates the need for iterative learning processes and streamlines the training of our urban flow prediction model. Meanwhile, it is important to note that urban flow data are commonly not recorded in a dense, continuous format due to real-world needs (e.g., buses have relatively large departure time intervals), rendering the dynamics of urban flow data not as microscopic or instantaneous (see Table \ref{tab:dataset_stats} for common urban flow intervals ranging from 30 minutes to 1 hour) as defined in continuous differentiable physics equations. Therefore, different from PGML tasks supported by fine-grained temporal data like weather or air pollution forecast \cite{zhan2022decomposition, jiang2020short}, using coarser urban flow datasets to fit a standard continuous model may in turn introduce unrealistically strong assumptions, hurting the model's real-world performance. 
In response, we discretize the dynamics of urban flow, and correspondingly develop a discrete Physics-guided Network (PN), which is simpler yet more effective compared with continuous methods such as \cite{ji2022stden}. Our experiments have further validated that the proposed discrete version of PN outperforms its continuous counterpart.

The main contributions of this paper are summarized below:
\begin{itemize}
    \item \textbf{New Problem and Insights.} We point out an unexplored problem of physical inconsistency of data when deploying PGML methods for urban flow prediction tasks. This new problem highlights the necessity of data-centric solutions that can effectively address these discrepancies, thereby enhancing both the robustness and capability of the model.
    
    \item \textbf{Novel Methodology.} We first propose a physics-guided network under the discrete setting which is a more natural fit for urban flow prediction scenarios. We then propose a framework for enhancing the capability of PN with the proposed active sample reweighting policy. The computed sample weights factor in both model uncertainty and physical consistency, thereby reinforcing the contribution of valuable data samples while alleviating the impact of low-quality ones during PN training.
    
    \item \textbf{State-of-the-art (SOTA) Performance.} We conduct extensive experiments on four real-world urban flow prediction datasets, and the experimental results indicate that our framework achieves SOTA performance against other baselines. In addition, our framework demonstrates strong robustness in experiments involving samples with varying levels of noise.
\end{itemize}

\section{Preliminaries}
\label{sec:definiton}
In this section, we first present the key concepts and definitions of continuity equations in urban flow graph, then formalize the research problem of this paper. \\

\noindent \textbf{Definition 1: Urban Grids and Urban Flow Graph.} The urban regions $\mathcal{V}=\{v_1,\dots,v_M\}$ within a city are partitioned into $H\times W$ grids. By connecting these spatial regions with edges we define an urban flow graph $\mathcal{G}=(\mathcal{V}, \mathcal{E}, \bm{x}_{1:T})$, where $\mathcal{E}$ is the set of edges connecting geographically adjacent regions, The urban flow data within these regions over a time period $T$ is denoted as $\bm{x}_{1:T} \in \mathbb{R}^{T \times M \times 2}$, of which the last dimension incorporates the inflow and outflow data for each region. Following common practices in urban flow modelling \cite{wang2019origin,ji2023spatio}, we can also represent the connectivity among the regions in $\mathcal{G}$ by using an adjacency matrix $\bm{A} \in \mathbb{R}^{M \times M}$.\\

\noindent \textbf{Definition 2: The General Continuity Equation.} Continuity equations are the fundamental physics used to describe movement and conservation laws of a particular quantity such as energy and electric charge \cite{lienhard2008doe}. The differential form of the general continuity equation can be defined as:
\begin{equation}
    \frac{\partial \bm{z}}{\partial t} + \nabla \cdot \bm{f} = 0,
\end{equation}
where $\bm{z}$ is the density of the quantity, $\bm{f}$ is the energy flux, $\nabla$ is the divergence operator, and $t$ is time.\\

\noindent \textbf{Definition 3: Urban Flow Continuity Equation.}
 We adopt the definition from \cite{ji2022stden} which posits that urban flow such as traffic can be described by the dynamics of urban energy fields. In \cite{ji2022stden}, the traffic energy flux is defined by the differences of traffic density based on the edge-wise traffic flow data. However, urban flow data is always recorded as node-wise data by the sensors in real scenarios \cite{xie2020urban}. By drawing upon the commonality with the energy flux and electric current \cite{landau2013classical}, we mildly assume that urban energy flux $\bm{f}$ is linearly related to the node-wise data of urban flows. Specifically, we define the urban energy flux in the urban energy fields as:
 \begin{equation}
     \bm{f} \approx -\bm{w}_s\nabla \bm{s}  + \bm{w}_r\nabla\bm{r},
     \label{eq:general_continuity_equaiton}
 \end{equation}
 where $\bm{w}_s$ and $\bm{w}_r$ are learnable parameters to compute the linear relationships between the urban energy flux and urban flow. $\bm{s}$ and $\bm{r}$ are the inflow and outflow, respectively. Particularly, $-(\nabla s_{ij})=-(s_i-s_j)$, and $\nabla r_{ij}=(r_i-r_j)$. It means that we utilize the discrete differences of urban flows between region nodes $i$ and $j$ to represent the graph gradients with different directions \cite{gustafson1985graph}, that is, the divergence of urban energy flux in the urban flow graph $\mathcal{G}$. By rewriting Eq.(\ref{eq:general_continuity_equaiton}), we define the urban flow continuity equation as a partial differential equation (PDE) form:
 \begin{equation}
 \begin{aligned}
 &\frac{\partial \bm{z}}{\partial t} + \nabla \cdot \bm{f} = 0, \\
     &\frac{\partial \bm{z}}{\partial t} - \nabla \cdot (\bm{w}_s\nabla \bm{s}  - \bm{w}_r\nabla\bm{r}) = 0,
 \end{aligned}
 \label{eq:urban_continuity_equaiton}
 \end{equation}
 where $\bm{z}=(z_1, \dots,z_N)$ is the potential density of the urban flow of each grid. Eq.(\ref{eq:urban_continuity_equaiton}) describes the principle of conservation between the urban energy flux and the potential density of urban flow in urban energy fields like the principles of conservation in fluid flows and electric charge \cite{barnard1966theory,schunk1975effect}. \\
 
\noindent \textbf{Problem Formalization.} Given the historical urban flow samples $\bm{x}_{1:T}$ and the urban flow graph $\mathcal{G}$, the task of this work is to accurately predict the future urban flow of each grid at $T+1$, denoted by $\bm{x}_{T+1} \in \mathbb{R}^{M\times 2}$.

\section{Methodology}
\begin{figure*}[h]
      \centering
      \includegraphics[width=0.9\linewidth]{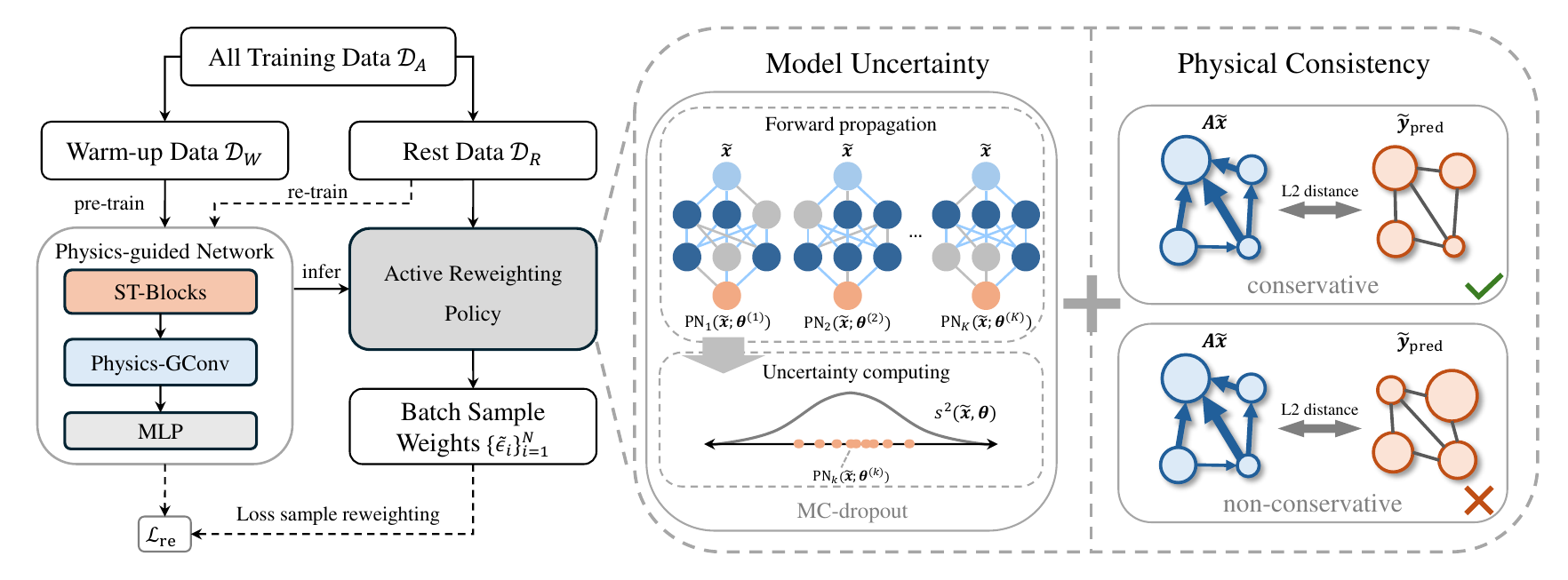}
      \caption{An overview of the P-GASR framework. Our P-GASR consists of two important parts: physics-guided network and active reweighting policy. In the proposed active reweighting policy, we design two components to compute the sample weights considering both data-driven and physics-guided aspects. We utilize Monte Carlo-dropout to measure model uncertainty, where $s^2(\tilde{\bm{x}},\bm{\Theta})$ represents the variance of $K$ prediction rounds. On the other hand, we use L2 distance between urban flow aggregation $\bm{A}\tilde{\bm{x}}$ and future prediction $\tilde{y}_{\text{pred}}$ to measure physical consistency based on the conservation law in discrete form.}
      \label{fig:P-GASR_overview}
      \vspace{-2mm}
    \end{figure*}

The Physics-guided Active Sample Reweighting (P-GASR) is a data-aware sample reweighting framework. As shown in Fig. \ref{fig:P-GASR_overview}, it mainly consists of two parts: physics-guided network and active reweighting policy. We provide further design details below.

\subsection{Physics-guided Network}
\label{sec:pe}
In this section, we propose a physics-guided network (PN) based on the urban flow continuity equation, which is shown in Eq. (\ref{eq:urban_continuity_equaiton}). As discussed earlier, training data and labels are recorded in intervals, i.e., each temporal data point represents a summary over a long period. Therefore, we alternatively focus on modelling and studying the dynamic system of urban flow in discrete form. To this end, we approximate Eq. (\ref{eq:urban_continuity_equaiton}) with discretized forward difference quotient:
\begin{equation}
    \frac{\bm{z}_{t+h}-\bm{z}_{t}}{h} = \nabla \cdot (\bm{w}_s\nabla \bm{s}_t  - \bm{w}_r\nabla\bm{r}_t),
    \label{eq:difference_quotient}
\end{equation}
where $h$ is the non-zero step size. Since the recorded time series data is sampled with uniform window sizes, we let $h=1$, and Eq. (\ref{eq:difference_quotient}) can be rewritten as:
\begin{equation}
    \bm{z}_{t+1} = \bm{z}_{t} +  (\bm{w}_s\Delta \bm{s}_t  - \bm{w}_r\Delta\bm{r}_t),
    \label{eq:discrete}
\end{equation}
where $\Delta \bm{x} = \nabla \cdot \nabla \bm{x}$ given $\bm{w}_s$ and $\bm{w}_r$ are irrelevant to $t$. We follow \cite{ji2022stden} to process the computation between Laplacian operator $\Delta$ and the representations of inflow and outflow $\bm{s}, \bm{r}$ using spectral graph neural network $g_{\bm{w}_x}(\bm{A},\bm{x})$, we then rewrite Eq. (\ref{eq:discrete}) as:

\begin{equation}
    \bm{z}_{T+1} = {\bm{z}}_{T} +  \sigma(g_{\bm{w}_s}(\bm{A},\bm{s}_T)) - \sigma(g_{\bm{w}_r}(\bm{A},\bm{r}_T)),
    \label{eq:pe}
\end{equation}
where $\sigma$ is the activation function. The derived PN in Eq. (\ref{eq:pe}) is a discrete form of residual graph neural network developed from the method in \cite{ji2022stden}. Previous works have proved that the mechanism of the residual network is effective for time series prediction \cite{zhang2017deep, wang2017time}. In Sec. \ref{sec:discussion}, we will discuss the difference between these two models from the technical aspect.

The relationship between urban flow $\bm{x}$ and potential urban density $\bm{z}$ can be represented by $\bm{x} = \bm{z} \cdot \bm{v}$ \cite{treiber2013traffic}, where $\bm{v}$ is the potential traffic speed. Given that traffic speed is normally unavailable in urban flow data, we alternatively utilize a learnable approach to learn the density $\bm{z}$ given the linear relationship between urban flow and density. Concretely, we learn the potential density representation ${\bm{z}}_{T}$ with spatio-temporal blocks (ST-Blocks):
\begin{equation}
    {\bm{z}}_{T} = \text{ST-Blocks}(\bm{s}_{1:T}, \bm{r}_{1:T}, \bm{s}_{1:T}-\bm{r}_{1:T}),
\end{equation}
where $\bm{s}_{1:T}$ is the input inflow's historical sample, $\bm{r}_{1:T}$ is the input outflow's historical sample. As aforementioned, the urban flows affect the potential density, thus we input the difference of inflow and outflow as additional guidance to learn the potential density.
In this study, we adopt the spatio-temporal encoder based on the combination of graph neural networks and temporal convolution networks from \cite{ji2023spatio} as the ST-Blocks. 

Finally, we decode the potential density to predict future urban flow $\hat{y}_{T+1}$ via a multilayer perceptron (MLP):
\begin{equation}
    \hat{y}_{T+1}=\text{MLP}({\bm{z}}_{T+1}),
    \label{eq:mlp}
\end{equation}
where the prediction is made for every grid's calculated density denoted by $ {\bm{z}}_{{T+1},i} \text{ for } i = 1, 2, ..., N.$

\subsection{Active Reweighting Policy}

Physics-guided models are able to address the issue of inaccurate predictions caused by data scarcity in data-driven models by leveraging principled domain knowledge. However, limited data availability implies that the observed data with low-quality samples is not fully consistent with physical principles, thereby limiting the predictive performance of the model, which can even be inferior to that of data-driven models. To address this limitation, we utilize sample reweighting based on the idea of active learning \cite{ren2021survey}. Unlike other sample reweighting methods, our approach does not require learning the sample weights for optimization such as meta-learning \cite{he2023physics,ren2018learning}. Instead, we utilize active reweighting policy (ARP), which is inspired by active learning policy to calculate the sample weights in a one-off manner during the test-time inference process. Furthermore, our ARP not only considers model uncertainty to enhance the robustness, but also incorporates physical knowledge in the computation of sample weights.
\\

\noindent \textbf{Model Optimization.} There are two training phases in P-GASR: pretraining and retraining. Different from the mechanism of pretraining and fine-tuning in transfer learning \cite{pan2009survey}, we pretrain and retrain different initialized models for computing sample weights. In this section, we will focus on introducing the optimization of the retrained PN and the computations involved in sample reweighting. In the retraining phase, the loss function for model optimization is:
\begin{equation}
    \mathcal{L}_\text{re}=\sum_{i=1}^N \tilde{\epsilon}_i \cdot(\lambda\left|y_{T+1, i}^{\text{in}}-\hat{y}_{T+1, i}^{\text{in}}\right|+(1-\lambda)\left|y_{T+1, i}^{\text{out}}-\hat{y}_{T+1, i}^{\text{out}}\right|),
    \label{eq:loss}
\end{equation}
where $y_{T+1, i}$ is the ground truth of urban flow, and $\lambda$ is the parameter for balancing the prediction of inflow and outflow. \\

\noindent \textbf{Sample Weight.}  In Eq.(\ref{eq:loss}), $\tilde{\epsilon}_i$ is the normalized sample weight in a mini-batch with size $N$:

\begin{equation}
   \tilde{\epsilon}_i = \frac{\exp \left(\epsilon_i\right)}{\sum_n^N \exp \left(\epsilon_n\right)} + b,
   \label{eq:norm_sample_weight}
\end{equation}
where $b$ is a constant for sample weight smoothing, avoiding any potential bias in the batch loss, which may arise when the majority of the sample weights in the batch are close to $0$. Here, $b=\frac{1}{N}$. 

The proposed ARP integrates both model uncertainty inherent in data-driven components and physical consistency essential in physics-guided components to reweight samples. Model uncertainty facilitates the quantification of uncertainty between predictions and observed data, thereby enhancing the model's robustness. Physical consistency directly improves the contribution of samples that conform to physical principles. Concretely, the calculation of sample weight ${\epsilon}_i$ in ARP is:
\begin{equation}
    {\epsilon}_i = \alpha u_i + \beta{c_i},
    \label{eq:sample_weight}
\end{equation}
where $u_i$ is model uncertainty, and $c_i$ is the measurement of physical consistency. $\alpha$ and $\beta$ serve as hyperparameters to balance the trade-off between these two components. We elaborate on the quantification of $u_i$ and $c_i$ in the following. \\

\noindent \textbf{Model Uncertainty.} The sample weights are computed in the process of test-time inference, given the inference dataset $\mathcal{D}_R=\{(\tilde{\bm{x}_i}, \tilde{y}_{T+1,i})\}_{i=1}^N$, the model uncertainty $u_i$ can be computed based on the variance of these prediction outputs using Monte Carlo-dropout \cite{gal2016dropout}. Specifically, we quantify the model uncertainty by computing the variance of the test-time predictions from $K$ rounds dropouts with different model parameters:
\begin{equation}
        u_i =\frac{1}{K-1} \sum_{k=1}^K\left(\text{PN}_k\left(\tilde{\bm{x}}_i; \bm{\Theta}^{(k)}\right)-\frac{1}{K}\sum_{k=1}^K\text{PN}_k\left(\tilde{\bm{x}}_i; \bm{\Theta}^{(k)} \right)\right)^2,
    \label{eq:model_uncertainty}
\end{equation}
where $\text{PN}_k$ is the physics-guided network at $k$-th dropout round, and $\bm{\Theta}^{(k)}$ is the corresponding parameter of $\text{PN}_k$. \\

\noindent \textbf{Physics Consistency.} On the other hand, we  use the reciprocal of L2 distance between prediction $\tilde{y}_{T+1,i}$ flows and the aggregated flows $\bm{A} \tilde{\bm{x}}_{T,i}$ from other nodes at the previous time step to measure physical consistency $c_i$:
\begin{equation}
    c_i = \left(\tilde{y}_{T+1,i} - \bm{A}\tilde{\bm{x}}_{T,i} \right)^{-2}.
    \label{eq:physical_inconsistency}
\end{equation}

This distance represents the intuition of conversation law in urban density network in discrete form \cite{bressan2015conservation}. Intuitively, when the predicted urban flow $\tilde{y}_{T+1,i}$ of the target node exactly equals the aggregation of the urban flow $\bm{A} \tilde{\bm{x}}_{T,i}$ from other nodes, this indicates that the sample perfectly conform to the conservation law. With the effect of inverse in Eq. (\ref{eq:physical_inconsistency}), larger value of $c_i$ reflects high-conservative, indicating a higher level of physical consistency. 

\subsection{Overall Process of P-GASR}
\label{sec:overall_process}

The optimization process of P-GASR is shown in Algorithm \ref{alg:P-GASR}. There are three important steps in the P-GASR process: 1. $D$-fold pretraining; 2. Model inference and sample weight computation; 3. Retraining. In line 2, we first initialize $D$ PNs which are defined in Sec. \ref{sec:pe}, for $D$-fold pretraining and $\text{PN}_{\text{re}}$ for retraining after deriving all sample weights $\tilde{\bm{\epsilon}}$. In lines 3-12, we conduct $D$-fold pretraining using Warm-up data $\mathcal{D}_W$ and compute $D$ parts of sample weights $\{\tilde{\bm{\epsilon}}_d\}_{d=1}^D$. After that, we retrain with $\text{PN}_{\text{re}}$ on all training data with sample weights for optimizing the loss function $\mathcal{L}_\text{re}$ (lines 14-15). We design the pretraining process inspired by the $K$-fold cross-validation process. This strategy ensures that the sample weight of each part can be calculated and enables the model to perform inference without requiring additional validation sets, as the sample weights are computed by using the remaining part of pretraining data after inference.

\begin{algorithm}
        \caption{Overall process of P-GASR.}
        \label{alg:P-GASR}
        \LinesNumbered
        \KwIn{All training data $\mathcal{D}_A$; adjacency matrix $\bm{A}$; hyperparameters $\alpha, \beta$ for balancing model uncertainty and physical consistency.} 
  
        \KwOut{Optimal model parameters.} 
        
        Split all training dataset $\mathcal{D}_A$ into $D$ parts $\{\mathcal{D}_1, \mathcal{D}_2, \dots, \mathcal{D}_D\}$;

        Initialize $\{\text{PN}_1, \text{PN}_2, \dots, \text{PN}_D\}$ and $\text{PN}_\text{re}$;
        
            \For{$d=1,\dots,D$}{
               \tcp{Pretrain.}
               Set Warm-up data $\mathcal{D}_W=\{\mathcal{D}_1, \mathcal{D}_2, \dots, \mathcal{D}_D\} -\mathcal{D}_d$;

                Pretrain $\text{PN}_{d}$ on $\mathcal{D}_W$: $\bm{\Theta}_d \leftarrow \text{PN}_d(\mathcal{D}_W)$, optimize $\text{PN}_{d}$ w.r.t. $\mathcal{L}_\text{p}=\sum_{i=1}^N (\lambda\left|y_{T+1, i}^{\text{in}}-\hat{y}_{T+1, i}^{\text{in}}\right|+(1-\lambda)\left|y_{T+1, i}^{\text{out}}-\hat{y}_{T+1, i}^{\text{out}}\right|)$;

                \tcp{Inference and compute sample weights.}
                Set Rest data $\mathcal{D}_R = \mathcal{D}_d$; 

                Infer PN on $\mathcal{D}_R$: $\tilde{\bm{y}}_{T+1, d}=\text{PN}_d(\mathcal{D}_R; \bm{\Theta}_d)$;

                Compute model uncertainty $\bm{u}_d=(u_i)_{i=1}^N$ with Eq. (\ref{eq:model_uncertainty});

                Compute physical consistency $\bm{c}_d=(c_i)_{i=1}^N$ with Eq. (\ref{eq:physical_inconsistency});

                Compute sample weight $\bm{\epsilon}_d=(\epsilon_i)_{i=1}^N$ with Eq. (\ref{eq:sample_weight});

                Normalize $\bm{\epsilon}_d$ to $\tilde{\bm{\epsilon}}_d$ with Eq. (\ref{eq:norm_sample_weight});
               
            }

            Combine sample weights from different parts $\bm{\tilde{\epsilon}} = \{\tilde{\bm{\epsilon}}_d\}_{d=1}^D$;

            \tcp{Retrain.}
            
            Retrain PN on all training data: $\hat{y}_{T+1}=\text{PN}_\text{re}(\mathcal{D}_A )$, optimize loss function w.r.t. Eq. (\ref{eq:loss}) using $\tilde{\bm{\epsilon}}$;
            
            \Return Optimal $\text{PN}_\text{re}$.
	\end{algorithm}

\subsection{Discussions}
\label{sec:discussion}
\subsubsection{Comparison between Our Discretized Physics-guided Network and STDEN \cite{ji2022stden}}
When developing PGML models, a default treatment is to formulate the system with neural ordinary differential equations (ODEs) that are continuous. In this regard, STDEN \cite{ji2022stden} is a relevant work on traffic forecasting, which uses ODE solver \cite{chen2018neural} to model the continuous density state. Our descretized PN, though bearing a similar physics-guided ethos, has a major advantage compared with STDEN. Firstly, STDEN is designed for predicting traffic as edge weights in transportation graphs, where each edge is a specific road connecting two sensor nodes \cite{ji2022stden}. 
In contrast, in general urban flow prediction tasks like passenger demand prediction for taxi services, nodes are geographic grids and edges only implicitly indicate their spatial affinity. As such, the urban flow is mostly recorded as node attributes, i.e., the quantity changes of vehicles/passengers of each grid in a time period. Secondly, unlike road traffic prediction that only concerns how many vehicles have been recorded on each road, predicting both the inflow and outflow for each grid is a unique yet indispensable aspect in urban flow prediction. Thus, the distinct problem setting/granularity and non-interchangeable data structures between two tasks render STDEN inapplicable to the urban flow prediction setting. To address these challenges, we redefine the urban flow continuity equation for modelling the urban flow in a potential energy field. In addition, our PN is a discrete approximation of the neural ODE, which aligns better with the nature of urban flow data. To compare the effect between the continuous and discrete modelling of urban flow dynamics, we develop a continuous version of PN (PN-con) to approximate the mechanism of STDEN as closely as possible, using the ODE solver to predict future potential density ${\bm{z}}_{T+1}$:
\begin{equation}
    {\bm{z}}_{T+1} = \text{ODEsolver}(g, {\bm{z}}_{T_0},[1,\dots,T]),
\end{equation}
where $g$ is the graph neural network similar in Eq. (\ref{eq:pe}), ${\bm{z}}_{T_0}$ is the initial state learned by ST-Blocks, and we decode the final density state by an MLP to predict the future flow $\hat{y}_{T+1}$ as the same as Eq. (\ref{eq:mlp}). We report the result of PN-con in Sec. \ref{sec:overall_comparison}. As previously discussed, the dynamics of urban flow are neither microscopic nor instantaneous, which leads to suboptimal performance from PN-con. Consequently, we develop a discrete version of PN which demonstrates better performance in the experiment.

\subsubsection{Difference between P-GASR and General Active Learning} 
The learning strategy of P-GASR is inspired by the deep active learning (DAL) strategy \cite{ren2021survey}. DAL aims to effectively select the most valuable data samples for labeling and training. By using the selected samples, this strategy aims to design a policy for achieving eligible performance while minimizing the amount of data that needs to be labeled and the associated costs. In essence, the core of DAL is effective data sampling. However, the purpose of P-GASR is not for data sampling, but rather to utilize the method of sample reweighting. By implementing the well-designed active reweighting policy, we aim to compute the sample weights, allowing data that conforms to physical principles to have a greater contribution to model training, rather than reducing the size of data. The integration of the DAL strategy with our P-GASR offers a significant advantage: unlike other sample reweighting methods that rely on time-consuming meta-learning to determine sample weights \cite{ren2018learning, shu2019meta}, P-GASR computes these weights directly through a single inference process.

\section{Experiments}
In this section, we design extensive experiments on four real-world urban flow prediction datasets to verify the effectiveness of P-GASR by answering the following research questions (RQ): 
\begin{itemize}
    \item \textbf{RQ1}: How does the proposed P-GASR perform compared to other state-of-the-art methods in urban flow prediction?
    \item \textbf{RQ2}: How does the active sample reweighting effect to different levels of data quality?
    
    \item \textbf{RQ3}: How do the components contribute to the performance of P-GASR?
    
    \item \textbf{RQ4}: How do the hyperparameters influence P-GASR?

\end{itemize}

\subsection{Experimental Settings}
\label{sec:experiment_set}
\subsubsection{Datasets}
We evaluate our proposed P-GASR on four widely used real-world urban flow datasets\footnote{https://github.com/Echo-Ji/ST-SSL\_Dataset} for passenger demand prediction: NYCBike1 \cite{zhang2017deep}, NYCBike2 \cite{yao2019revisiting}, NYCTaxi \cite{yao2019revisiting} and BJTaxi \cite{zhang2017deep}, the statistics of these datasets are summarized in Table \ref{tab:dataset_stats}. 

We follow all the settings of \cite{ji2023spatio}, using the four-time steps before the current time step, the same time step three days before and two-time steps before and after the current time step to predict the next future step. We also use the sliding windows to generate samples, we then split the dataset into a training set, validation set and test set with a ratio of 7:1:2.

\begin{table}[h]
          \caption{Statistics of four urban flow prediction datasets. }
          \label{tab:dataset_stats}
          \renewcommand{\arraystretch}{0.9}
          \resizebox{0.48\textwidth}{!}{
          \begin{tabular}{c|cccc}
            \toprule
            Dataset & Interval & \# Regions & \# Vehicles & Time Span \\
            \midrule
            NYCBike1 & 1 hour & 16$\times$8 & 6.8k+ & 04/01/2014-09/30/2014\\
            NYCBike2 & 30 min & 10$\times$20 & 2.6m+ & 07/01/2016-08/29/2016\\
            NYCTaxi & 30 min & 10$\times$20 & 22m+ & 01/01/2015-03/01/2015\\
            BJTaxi & 30 min & 32$\times$32 & 34k+ & 03/01/2015-06/30/2015\\
            \bottomrule
          \end{tabular}}
    \end{table}

    \begin{table*}[t]
    \centering
     \caption{Overall performance comparison of models in terms of inflow and outflow on four datasets. The best results are highlighted in bold, and the second-best results are highlighted with underlines.}
    \setlength{\tabcolsep}{4pt}
    \renewcommand{\arraystretch}{0.9}{
      \begin{tabular}{c||cc|cccccc|cc|c}
      \toprule
        Dataset    & \multicolumn{1}{c|}{Metric} & Type  & ARIMA & SVR   & ST-ResNet & STGCN & STID  & ST-SSL & \textbf{PN-con} & \textbf{PN-dis} & \textbf{P-GASR} \\
      \hline
      \midrule
    \multirow{4}{*}{\rotatebox{0}{NYCBike1}} & \multicolumn{1}{c|}{\multirow{2}{*}{MAE}} & In    & 10.66  & 7.27  & 5.36±0.05 & 5.02±0.04 & 5.30±0.03 & 4.96±0.03 & 4.92±0.05 & \underline{4.90±0.02}  & \textbf{4.88±0.02} \\
        & \multicolumn{1}{c|}{} & Out   & 11.33  & 7.98  & 5.66±0.06 & 5.29±0.05 & 5.58±0.02 & 5.27±0.03 & 5.20±0.05 & \underline{5.19±0.03} & \textbf{5.17±0.01} \\
        \cline{2-12}        & \multicolumn{1}{c|}{\multirow{2}{*}{MAPE}} & In    & 33.05  & 25.39  & 25.39±0.22 & 24.03±0.57 & 25.66±0.09 & 23.48±0.27 & 23.43±0.20 & \underline{23.35±0.14} & \textbf{23.33±0.21} \\
        & \multicolumn{1}{c|}{} & Out   & 35.03  & 27.42  & 26.25±0.22 & 24.98±0.25 & 26.14±0.12 & 24.37±0.49 & \underline{24.02±0.26} & 24.03±0.31 & \textbf{23.95±0.05} \\
      \midrule
      \midrule
      \multirow{4}{*}{\rotatebox{0}{NYCBike2}} & \multicolumn{1}{c|}{\multirow{2}{*}{MAE}} & In    & 8.91  & 12.82  & 5.54±0.11 & 5.26±0.03 & 5.24±0.04 & 5.06±0.04 & 5.25±0.10 & \textbf{5.04±0.02} & \underline{5.05±0.01} \\
          & \multicolumn{1}{c|}{} & Out   & 8.70  & 11.48  & 5.19±0.09 & 4.91±0.03 & 4.92±0.04 & 4.75±0.03 & 4.88±0.11 & \underline{4.69±0.02}  & \textbf{4.68±0.02} \\
          \cline{2-12}         & \multicolumn{1}{c|}{\multirow{2}{*}{MAPE}} & In    & 28.86  & 46.52  & 25.09±0.35 & 23.16±0.46 & 22.86±0.11 & 22.70±0.34 & 22.81±0.46 & \underline{22.38±0.44} & \textbf{22.15±0.07} \\
          & \multicolumn{1}{c|}{} & Out   & 28.22  & 41.91  & 24.04±0.25 & 22.00±0.42 & 21.92±0.19 & 21.53±0.39 & 21.43±0.13 & \underline{21.42±0.25}  & \textbf{21.06±0.27} \\
\midrule
\midrule
    \multirow{4}{*}{\rotatebox{0}{NYCTaxi}} & \multicolumn{1}{c|}{\multirow{2}{*}{MAE}} & In    & 20.86  & 52.16  & 13.99±0.29 & 12.43±0.20 & 12.35±0.08 & 12.37±0.31 & 12.23±0.03 & \underline{12.05±0.06} & \textbf{11.92±0.04} \\
          & \multicolumn{1}{c|}{} & Out   & 16.80  & 41.71  & 11.24±0.31 & 10.31±0.28 & 10.13±0.05 & 10.08±0.24 & 10.09±0.02 & \underline{9.82±0.09} & \textbf{9.69±0.08} \\
          \cline{2-12}          & \multicolumn{1}{c|}{\multirow{2}{*}{MAPE}} & In    & 21.49  & 65.10  & 20.53±0.46 & 18.27±0.81 & 17.13±0.21 & 17.49±0.50 & 17.05±0.10 & \underline{16.94±0.71} & \textbf{16.42±0.22} \\
          & \multicolumn{1}{c|}{} & Out   & 21.23  & 64.06  & 20.60±0.70 & 20.28±0.81 & 17.99±0.40 & 17.65±0.23 & \underline{16.89±0.26} & 17.38±0.35 & \textbf{16.76±0.44} \\
\midrule
\midrule
    \multirow{4}{*}{\rotatebox{0}{BJTaxi}} & \multicolumn{1}{c|}{\multirow{2}{*}{MAE}} & In    & 21.48  & 52.77  & 12.39±0.02 & 11.41±0.06 & 11.95±0.10 & 11.37±0.02 & 11.66±0.13 & \underline{11.32±0.04} & \textbf{11.25±0.08} \\
          & \multicolumn{1}{c|}{} & Out   & 21.60  & 52.74  & 12.49±0.02 & 11.50±0.06 & 11.98±0.09 & 11.46±0.03 & 11.75±0.12 & \underline{11.41±0.06} & \textbf{11.34±0.08} \\
\cline{2-12}          & \multicolumn{1}{c|}{\multirow{2}{*}{MAPE}} & In    & 23.12  & 65.51  & 17.56±0.15 & 15.99±0.44 & 15.96±0.36 & \underline{15.49±0.52} & 16.64±1.53 & 15.69±0.25 & \textbf{15.24±0.22} \\
          & \multicolumn{1}{c|}{} & Out   & 20.67  & 65.51  & 17.72±0.15 & 16.16±0.51 & 16.39±0.82 & \underline{15.55±0.43} & 16.60±1.42 & 15.86±0.26 & \textbf{15.35±0.23} \\
    \bottomrule
    \end{tabular}}%
    \label{tab:overall_comparison}%
  \end{table*}%

\subsubsection{Baselines}
We show the superiority by comparing 8 baselines in urban flow prediction. The details of these baselines are as follows:
\begin{itemize}
    \item \textbf{ARIMA} \cite{kumar2015short} is a moving average-based autoregressive model.
    \item \textbf{SVR} \cite{castro2009online} is the support vector machine for regression tasks.
    
    \item \textbf{ST-ResNet} \cite{zhang2017deep} is a convolution network-based model with residual function for spatio-temporal urban flow prediction.
    
    \item \textbf{STGCN} \cite{yu2017spatio} is a graph convolutional network integrates 1D convolution network to model spatio-temporal data.

    \item \textbf{STID} \cite{shao2022spatial} is a simple MLP-based model attaching spatial and temporal identity information.
    
    \item \textbf{ST-SSL} \cite{ji2023spatio} is the SOTA in urban flow prediction integrating spatial and temporal self-supervised learning.
    
    \item \textbf{PN-dis} is the discrete version of the physics-guided network, which is introduced in Sec. \ref{sec:pe}. 
    
    \item \textbf{PN-con} is the continuous version of the physics-guided network training with Eq.(\ref{eq:urban_continuity_equaiton}) and ODE solver \cite{chen2018neural}.
\end{itemize}

\subsubsection{Evaluation Metrics and Implementation Details}
We use two common metrics for evaluating the performance of models, which are Mean Average Error (MAE) and Mean Average Percentage Error (MAPE). For a fair comparison, we train each model four times. We report the average inflow and outflow results and the standard deviations in the experiment of overall comparison and report the average of inflow and outflow in the rest of the experiments.

We implement our P-GASR in Pytorch and conduct the experiments using RTX 4090 GPU. We set the learning rate at 0.001 and use the Adam optimizer for training. The batch size is 20 for the BJTaxi dataset and 32 for others, and the dimension of embedding is 64. Besides, we adopt the early-stop strategy, and we set the stop patience of pretrain models at 15 and the retrain model at 30. The number of inference runs $K$ for MC-dropout is 10. We adopt ChebNet \cite{defferrard2016convolutional} as the graph neural network in Eq. (\ref{eq:pe}). The activation function used in the graph neural network is the ReLU function, and the MLP uses the Tanh function.
\begin{figure*}[h]
      \centering
      \includegraphics[width=.83\linewidth]{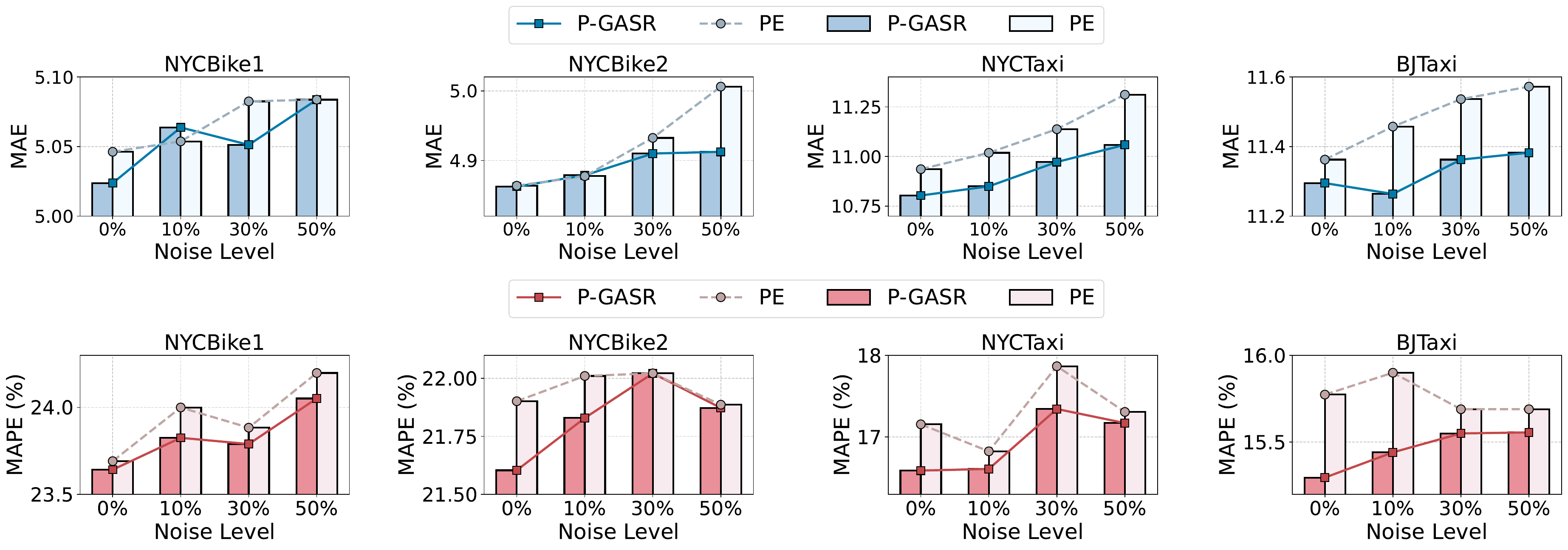}
      \caption{Influence of different noise levels to P-GASR and PN.}
      \label{fig:noise_influence}
 \end{figure*}
 
 \begin{figure*}[h]
      \centering
      \includegraphics[width=0.82\linewidth]{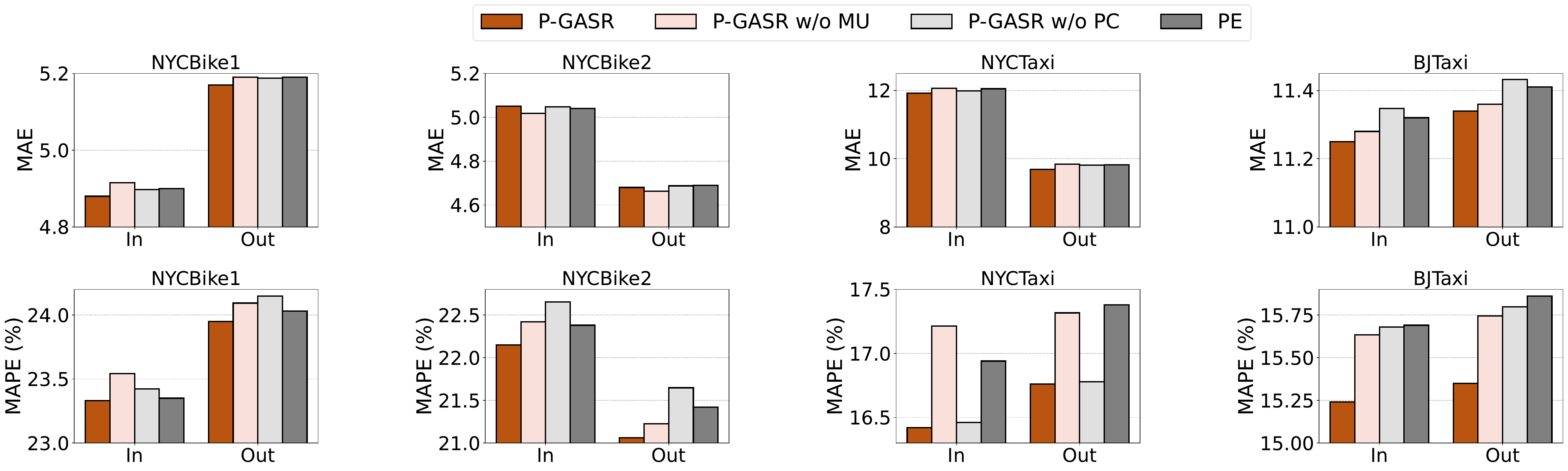}
      \caption{Performance comparison of ablation study with the variants of P-GASR.}
      \label{fig:ablation_study}
      \vspace{-2mm}
 \end{figure*}

\subsection{Overall Performance Comparison (RQ1)}
\label{sec:overall_comparison}
We compare our P-GASR with 8 baselines on NYCBike1, NYCBike2, NYCTaxi and BJTaxi. Note that since our setting is totally consistent with \cite{ji2023spatio}, and there is no randomness in ARIMA and SVR, we directly use the results of these two models from \cite{ji2023spatio}. The overall comparison result is shown in Table \ref{tab:overall_comparison}. 

From the overall comparison result, we can draw the following conclusions: (1) The proposed P-GASR achieves the best results on four datasets. The proposed physics-guided active sample reweighting mechanism alleviates the problem of physical inconsistency, successfully improving the result of PN. (2) The discrete version PN outperforms other baselines on NYCBike2 and NYCTaxi, including the SOTA ST-SSL, with improvements up to 3.2\%. This is attributed to the effectiveness of its physics-guided component compared to ST-SSL, which uses the same ST-Blocks. (3) The discrete version of PN surpasses the continuous version of PN. This result indicates that the continuous architecture performs worse in this case of urban flow prediction, but a simplified discrete architecture can precisely capture the pattern of urban flow.

Overall, the above results and observations indicate that P-GASR achieves improvements up to 6.1\% compared with the SOTA ST-SSL, and improves the performance of vanilla PN on four widely used datasets, which verifies its superiority in urban flow prediction and the effectiveness by alleviating the problem of physical inconsistency between the physics-guided model and training data.

\vspace{-2mm}

\subsection{Effectiveness of Active Sample Reweighting (RQ2)}
\label{sec:noise_influence}
The proposed active reweighting policy (ARP) aims to regularize the contribution of samples to alleviate physical inconsistency. We conduct the experiment to verify the effectiveness of sample weights by investigating the robustness of P-GASR and PN at different synthetic noise levels. Technically, we randomly replace standardized data samples with Gaussian noise in the levels of $[10\%, 30\%, 50\%]$ to artificially
reduce the quality of data. The result of the influence of different noise levels on our P-GASR and PN is shown in Fig. \ref{fig:noise_influence}. 

From the experimental result, we have the following observations: (1) PN performs worse than P-GASR at different noise levels. (2) The higher the noise level, the worse PN performs, while P-GASR shows less deterioration as the level increases. (3) PN might perform worse at low noise levels than at high noise levels, but the relationship between P-GASR and noise level is approximately linear. These observations prove that the noise generated by our proposed ARP can effectively mitigate the impact of noise samples, and as the noise level increases, the performance of P-GASR is not greatly affected, especially at high noise levels.

\subsection{Ablation Study (RQ3)}
To investigate the effectiveness of each component in P-GASR, we conduct an ablation study by comparing the performance with three variants. Here, we consider three variants: (1) P-GASR w/o MU removes the part of model uncertainty in the sample weight computation. (2) P-GASR w/o PC removes the part of physical consistency in the sample weight computation. (3) PN is the discrete version of a physics-guided network and can be considered as a variant of P-GASR where all sample weights are the same. 

The result of the ablation study is shown in Fig. \ref{fig:ablation_study}. From the result, we can observe that P-GASR outperforms its variants, which means both contributions of model uncertainty and physical consistency are significant for sample reweighting. The variants of P-GASR w/o MU and P-GASR w/o PC have different performances in different datasets, the performance of P-GASR w/o MU indicates that model uncertainty is more significant on NYCBike1 and NYCTaxi. On the contrary, Physical consistency contributes to improving the performances on NYCBike2 and BJTaxi. In addition, these two variants are even worse than PN in some cases.


\begin{figure*}[t]
    \centering
    \begin{minipage}[b]{.325\linewidth}
        \centering
        \includegraphics[width=\linewidth]{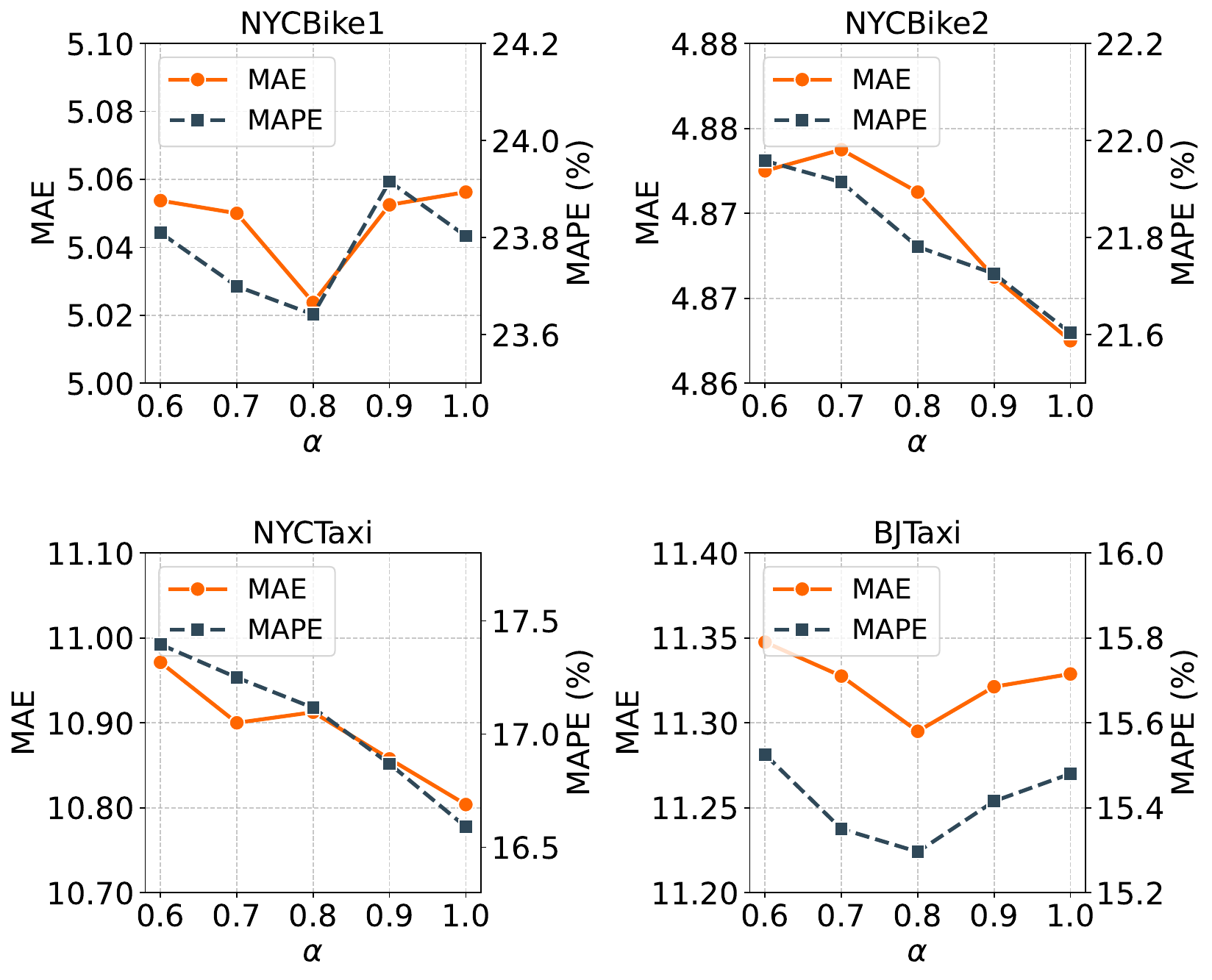}
        \subcaption{Influence of model uncertainty parameter $\alpha$ to P-GASR.}
        \label{fig:param_alpha}
    \end{minipage}\hfill
    \begin{minipage}[b]{.325\linewidth}
        \centering
        \includegraphics[width=\linewidth]{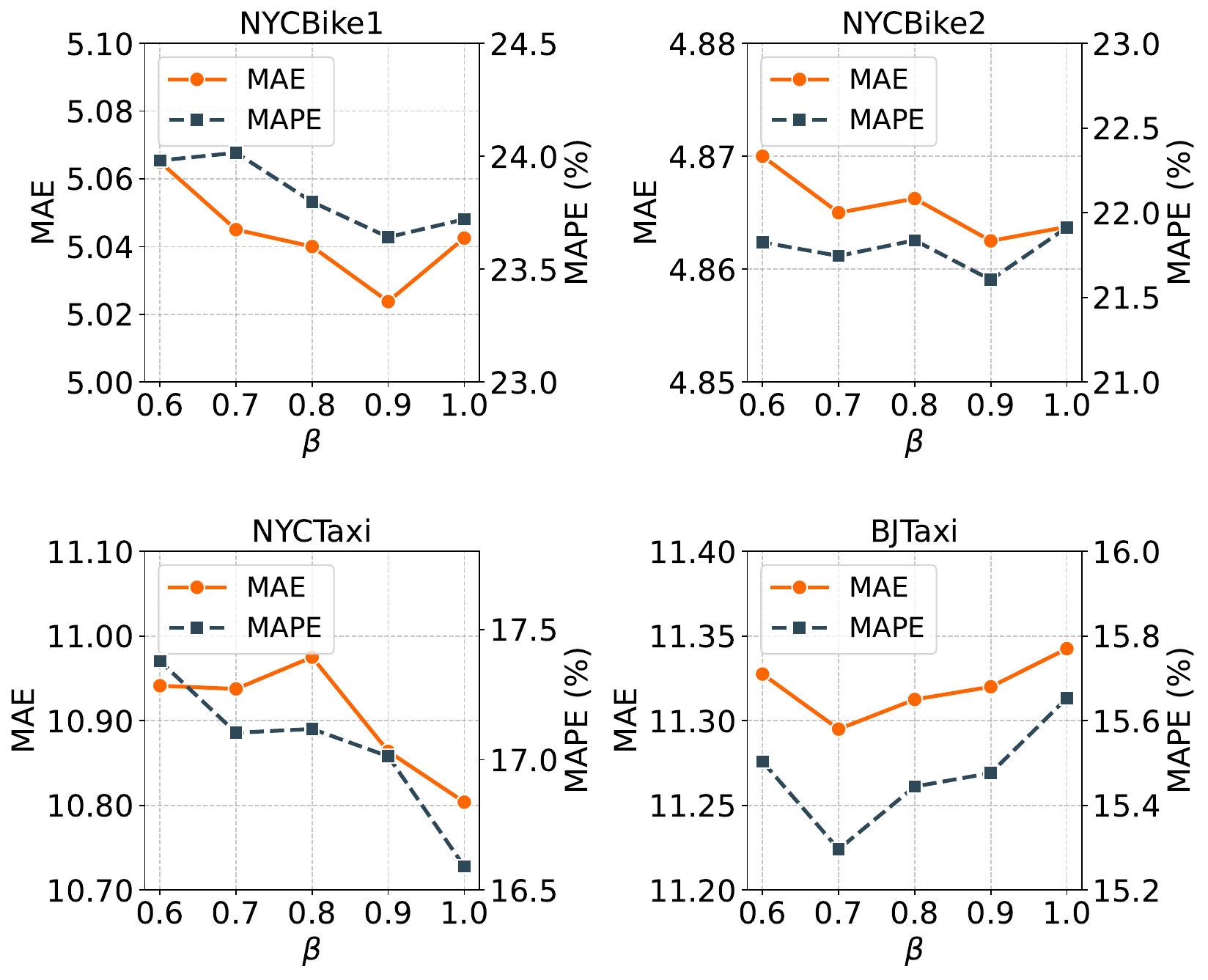}
        \subcaption{Influence of physical consistency parameter $\beta$ to P-GASR.}
        \label{fig:param_beta}
    \end{minipage}\hfill
    \begin{minipage}[b]{.325\linewidth}
        \centering
        \includegraphics[width=\linewidth]{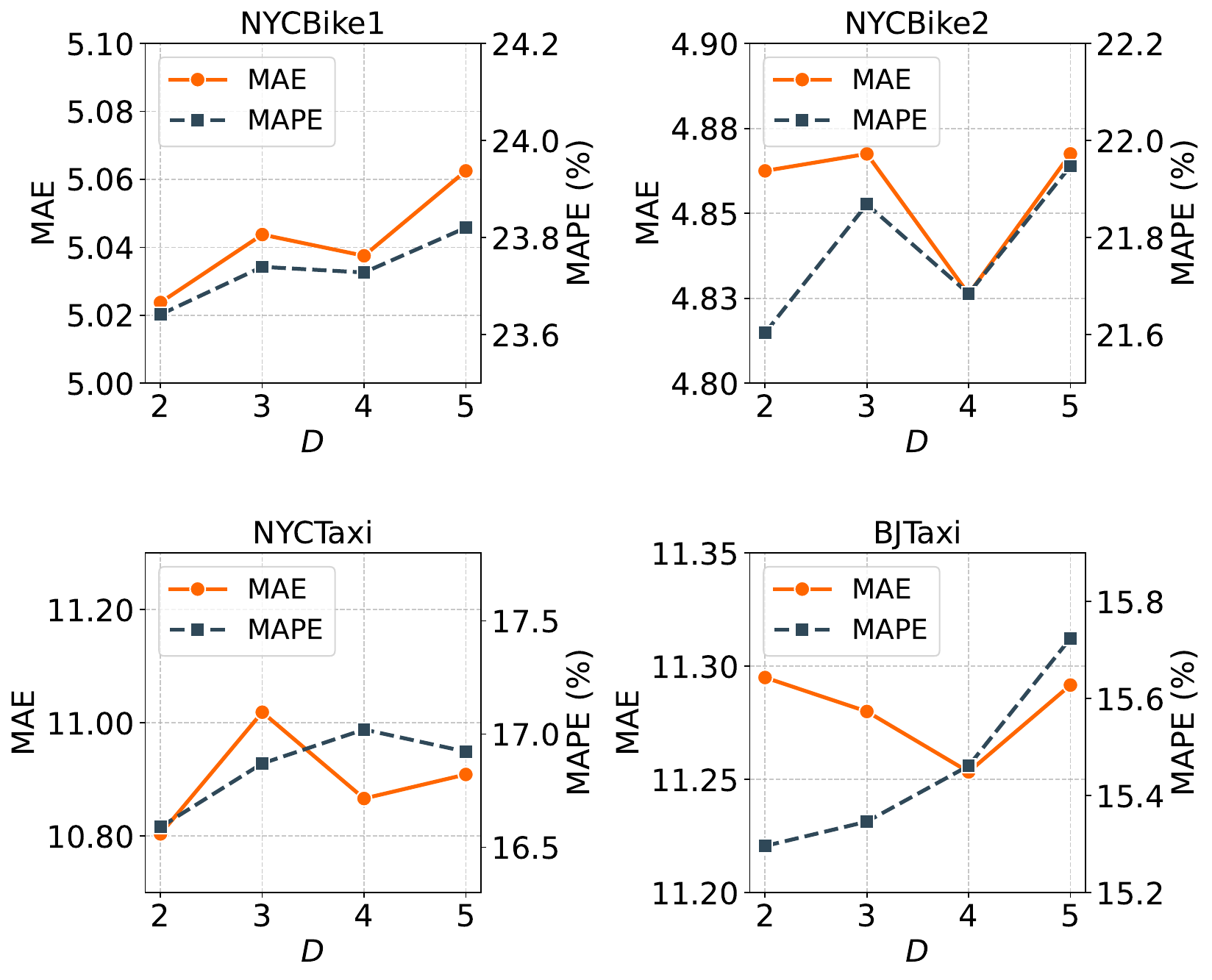}
        \subcaption{Influence of the number of pretraining runs $D$ to P-GASR.}
        \label{fig:param_d}
    \end{minipage}
    \caption{Influence of hyperparameters to P-GASR on four passenger demand datasets.}
    \label{fig:param}
    
\end{figure*}

\subsection{Hyperparameter Analysis (RQ4)}
In this section, we conduct the experiment to investigate the influence of three key hyperparameters in P-GASR: (1) $\alpha$ is the parameter to control the contribution of model uncertainty. (2) $\beta$ is the parameter to control the contribution of physics consistency. (3) $d$ is the parameter to set the number of folds for pre-training. Particularly, we set $\alpha$ and $\beta$ in the range of $\{0.6,0.7,0.8,0.9,1.0\}$, and set $d=\{2,3,4,5\}$. To investigate the target hyperparameter, we fix other hyperparameters by default.

The results of hyperparameter analysis are shown in Fig. \ref{fig:param}, the result indicates that P-GASR achieves the best performances at $\alpha=0.8$ on NYCBike1 and BJTaxi, and achieves the best performances at $\alpha=1$ on NYCBike2 and NYCTaxi, and the performance is reducing when decreasing  $\alpha$. Besides, P-GASR can achieve the best performances at $\beta=0.9$ on NYCBike1 and NYCBike2, and can be the best at $\beta=1$ and $\beta=0.7$ on NYCTaxi and BJTaxi, respectively. These results suggest that our model is influenced by $\alpha$ and $\beta$ to a certain extent. However, the performance of P-GASR is not influenced significantly by $D$. Increasing $D$ leads to higher time consumption in our method, therefore, we set $D=2$ on all datasets. 

\section{Related Work}
\subsection{Urban Flow Prediction}
Urban flow prediction is crucial for enhancing urban planning and optimizing vehicle dispatching, the main challenge of this task is how to capture the features of spatial and temporal information. Time series models like ARIMA \cite{kumar2015short} and support vector regression \cite{castro2009online} are widely used for handling this task, however, these models are failed to capture the spatial features in grid datasets. To address this limitation, initial works use convolution neural network (CNN) to process grid data as images \cite{zhang2017deep, chai2018bike, yao2018modeling}. Besides, some works adopt recurrent neural networks (RNN) to improve the ability to capture temporal dependency \cite{wei2018residual, du2019deep, zheng2016keyword, wang2021secure}. After the rise of graph neural networks (GNN), many remarkable works integrating GNN and 1-dimensional CNN to model grid data by capturing spatial and temporal information simultaneously \cite{yu2017spatio,diao2019dynamic,ye2021coupled,ji2023spatio, nguyen2017argument, wang2019origin, wang2021passenger}. However, these methods focus on constructing a data-driven architecture, which neglects physical principles. Inspired by \cite{ji2022stden}, we propose physics-guided methods to address this limitation.

\subsection{Physics-guided Machine Learning}
To address the limitation of black-box structures, recent proposed machine learning methods integrate physical principles with data-driven models for time series applications in the fields of epidemic, meteorology, river flow, air pollution and crop yield \cite{rodriguez2023einns,he2023physics,hettige2024airphynet,chen2023physics,chen2022physics,jia2021physics}. 
These methods enhance model prediction by incorporating physical knowledge from various fields, such as fluid dynamics, aerodynamics, and the laws of conservation of mass and energy. Some methods address the issue of insufficient training data in data-driven models by utilizing physics-based models to simulate additional datasets under varying parameter settings, thereby enhancing the view of data \cite{chen2023physics,chen2022physics,jia2021physics}. Many studies claim that the learning effectiveness of data-driven models is limited due to sparse data, which prevents the models from capturing the underlying physical laws of the systems. To handle this, a common approach modifies the loss function by incorporating various physical laws, including initial conditions, boundary conditions, and other physical properties \cite{rodriguez2023einns,he2023physics}. These methods constrain the learning direction of the model without altering its structures or mechanisms, changing only the loss functions. With the rise of Neural Ordinary Differential Equations \cite{chen2018neural}, the time consumption of using neural networks to solve ordinary differential equations (ODEs) has significantly reduced. Consequently, many methods are not merely modifying the loss function. Instead, they define the transformations of ODEs to replace traditional graph convolutional structures \cite{ji2022stden, hettige2024airphynet}. These models employ ODE solvers to predict future states of physical variables, integrating a more dynamic modelling approach that directly incorporates the temporal dynamics of the systems.

In this study, we follow \cite{ji2022stden,hettige2024airphynet} and reconstruct the layer of GNN, and solve the problem of physical inconsistency by using the technique of sample reweighting. Particularly, He \textit{et al.} proposed a physics-guided method combining sample reweighting \cite{he2023physics}. This method focuses on solving the problem of domain shift using a meta-learning strategy, while we focus on solving the problem of physical inconsistency using an active learning strategy.


\section{Conclusion}
In this paper, we study the physical inconsistency problem of physics-guided machine learning methods in urban flow prediction. To overcome this problem, we propose a physics-guided active sample reweighting framework. Technically, we design an active reweighting policy integrating the measurements of model uncertainty and physical consistency to compute sample weights. In addition, we develop a discrete physics-guided network for modelling urban flow based on the continuity equation, and use our framework to enhance its capability. We verify the effectiveness of the proposed P-GASR by conducting a series of experiments on four real-world datasets. The experimental results indicate that the proposed P-GASR achieves state-of-the-art performance and alleviates the physical inconsistency problem.

\section{Acknowledgement}
This work is supported by Australian Research Council under the streams of Future Fellowship (Grant No. FT210100624), Linkage Project (Grant No. LP230200892), Discovery Early Career Researcher Award (Grants No. DE230101033), and Discovery Project (Grants No. DP240101108 and No. DP240101814).

\bibliographystyle{ACM-Reference-Format}
\balance
\bibliography{sample-base}


\end{document}